\documentclass[runningheads]{llncs}
\usepackage[T1]{fontenc}
\usepackage[numbers]{natbib}
\usepackage{graphicx}
\usepackage{verbatim}

\begin{document}
\title{Deepchecks: Evaluating Retrieval-Augmented Generation (RAG)}
%
%
\author{
    Assaf Gerner\inst{1} \and
    Netta Madvil\inst{1} \and
    Nadav Barak\inst{1} \and
    Alex Zaikman\inst{1} \and
    Jonatan Liberman\inst{1} \and
    Liron Hamra\inst{1} \and
    Rotem Brazilay\inst{1} \and
    Shay Tsadok\inst{1} \and
    Yaron Friedman\inst{1} \and
    Neal Harow\inst{1} \and
    Noam Bresler\inst{1} \and
    Shir Chorev\inst{1} \and
    Philip Tannor\inst{1} \and
    Lior Rokach\inst{2}
}
%
%
\institute{
    Deepchecks, Ramat Gan, Israel
    \and
    Ben-Gurion University, Beer Sheva, Israel
    \email{liorrk@bgu.ac.il}
}
\maketitle              
\pagestyle{plain}
\begin{abstract}
Large Language Models (LLMs) augmented with Retrieval-Augmented Generation (RAG) techniques are revolutionizing applications across multiple domains, such as healthcare, finance, and customer service. Despite their potential, evaluating RAG systems remains a complex challenge due to the stochastic nature of generated outputs and the intricate interplay between retrieval and generation components. This paper introduces Deepchecks, a comprehensive framework tailored for evaluating RAG applications. Deepchecks' evaluation framework addresses RAG applications evaluation through a multi-faceted approach, root cause analysis and production monitoring. By ensuring alignment with application-specific requirements, Deepchecks framework provides a robust foundation for assessing reliability, relevance, and user satisfaction in RAG systems.
\end{abstract}

\section{Introduction}
Retrieval-Augmented Generation (RAG) is a framework that enhances the capabilities of Large Language Models (LLMs) by integrating external knowledge sources into the generation process. By combining the text generation prowess of LLMs with information retrieval functions, RAG enables the production of precise and contextually relevant information. This approach addresses several critical limitations of traditional LLMs, such as their propensity to generate generic responses or false claims, often referred to as hallucinations.

The RAG process typically involves the following steps:
\begin{itemize}
    \item \textbf{External Data Curation:} External data from various sources, such as APIs, databases, or document repositories, is converted into numerical representations using embedding language models and stored in a vector database.
    \item \textbf{Information Retrieval:} Upon receiving a user query, the RAG system retrieves relevant information from the external knowledge base.
    \item \textbf{Augmented Generation:} The retrieved information is combined with the user query and fed into the LLM, which uses this augmented context along with its training data to generate a response.
\end{itemize}

\begin{figure}[ht]
\centering
\includegraphics[width=0.8\textwidth]{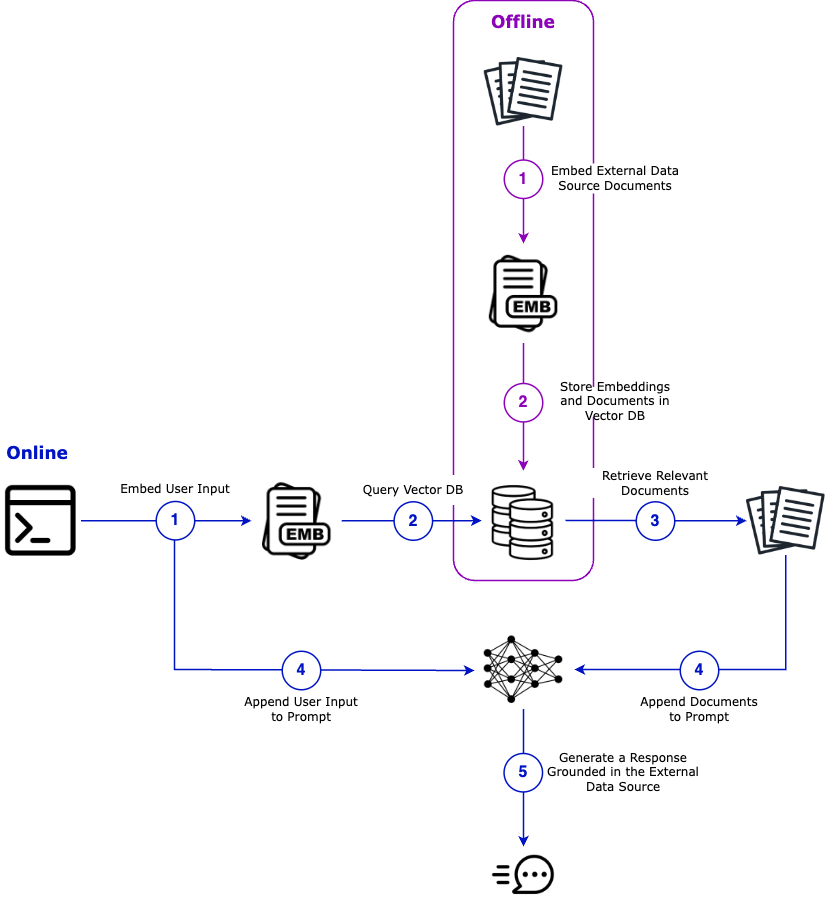}
\caption{: The core principles of a RAG system: (1) Offline – store external data, serving as a source of truth, in a vector database. (2) Online – augment the user query by retrieving and incorporating relevant documents from the database.}
\label{fig:rag-structure}
\end{figure}

Despite the promise of RAG systems, evaluating their performance remains a challenge. RAG systems comprise multiple interconnected components, each requiring unique evaluation methodologies. In turn, this complicates end-to-end assessment; i.e., providing a comprehensive unified scoring for RAG pipelines.

Addressing these challenges demands a multifaceted and reference-free approach that integrates automated metrics capable of capturing the unique characteristics of RAG systems. The evaluation of RAG applications has become a critical area of research, focusing on systematic methodologies to assess model capabilities across dimensions such as alignment, robustness, and domain-specific performance. Ensuring alignment with user needs, domain-specific requirements, and performance goals is essential for advancing this technology.

This paper explores the challenges associated with evaluating RAG applications and introduces the Deepchecks framework as a comprehensive solution, assisting evaluation, iterating system improvements, and production monitoring. By addressing key limitations in existing evaluation methodologies, Deepchecks offers a robust and modular approach to assess RAG pipelines effectively, ensuring their reliability and practical applicability.

\section{Related Work}
Evaluating Retrieval-Augmented Generation (RAG) systems has become a critical focus in modern NLP research, spurring the development of several frameworks designed to provide robust, scalable, and task-agnostic evaluation methodologies. This section reviews prior work, organizing the discussion around evaluation metrics and methods, domain-specific and general frameworks, and emerging techniques.

\subsection{Retrieval Evaluation}

\subsubsection{Classic Methods}
\hfill \break

Retrieval evaluation using classic methods relies on well-established metrics and benchmarks to assess both retrieval quality and generation fidelity.  For instance, retrieval metrics such as NDCG@k and MRR@k evaluate the ranking effectiveness of retrieved results. NDCG@k evaluates the ranking quality of retrieved results, rewarding highly relevant items appearing at the top. It is useful when multiple levels of relevance matter and the overall ranking order is important. MRR@k focuses on the position of the first relevant result, making it ideal for tasks where retrieving the correct answer quickly is crucial. While NDCG@k is preferred for assessing all retrieved results, MRR@k is best suited for scenarios where finding the first relevant result is the priority.

Effective evaluation also requires standardization in the form of robust benchmarks. For instance, the BEIR and MS-Marco benchmarks facilitate comprehensive assessment, standardization, and real-world applicability of retrieval-augmented generation systems. BEIR (Benchmarking IR) is a widely used benchmark for evaluating information retrieval tasks, BEIR provides a standardized framework with diverse datasets, supporting thorough testing across multiple domains. MS-Marco focuses on question-answering and ranking tasks, MS-Marco features large-scale real-world queries, enabling the evaluation of retrieval performance in practical settings. By leveraging these benchmarks, researchers can enhance the effectiveness and reliability of RAG systems.

\subsubsection{LLM-based evaluation}
\hfill \break

LLM-based evaluation, particularly the LLM-as-a-Judge approach, offers a powerful and flexible method for assessing retrieval due to its ability to capture nuanced aspects of textual data. The evaluator LLM is provided with specific criteria and prompted to assess various aspects of the retrieved content.

Several automated metrics powered by LLMs have emerged for specific retrieval evaluation dimensions. \textit{Context relevance} assessment measures the proportion of retrieved documents actually needed to answer the query, calculated through LLM-based sentence-level necessity analysis \cite{promptfoo2024rag}. \textit{Multimodal recall} evaluation, as implemented in \cite{zaheer2024multimodal}, uses vision-language models to assess product image-text alignment in e-commerce retrieval scenarios. For multilingual systems, \cite{zaheer2024multimodal} developed a framework where LLMs generate localized annotation guidelines before performing culture-aware relevance judgments.

Using LLMs as judges for evaluation comes with several limitations. Their assessments can be subjective, leading to variability and potential biases in judgment. Additionally, the computational cost of running multiple LLMs for evaluation is high, making large-scale assessments resource-intensive. Ensuring consistency across different prompts and models is another challenge, as variations in responses can lead to inconsistent evaluations, reducing the reliability of the assessment process. Studies show that without proper calibration, LLM judges can exhibit position bias favoring earlier documents in retrieval lists \cite{vespa2024retrieval}. Hybrid approaches combining LLM scoring with traditional metrics like BM25 term overlap help maintain grounding in lexical relevance \cite{genir2024}. Recent benchmarks indicate that GPT-4-based evaluation achieves 0.81 Spearman correlation with human judgments when using chain-of-thought prompting for retrieval assessment \cite{confident2024judge}.

\subsection{Generation Evaluation}
\subsubsection{Reference-based Evaluation}
\hfill \break

For assessing generated responses, BLEU, ROUGE, and METEOR are commonly used to compare generated text with reference outputs, capturing lexical overlap and coherence. However, these classic methods have limitations when applied to RAG systems, as they often fail to account for contextual alignment, factuality, and nuanced semantic correctness, necessitating the development of more specialized evaluation frameworks.

BERTScore is a neural-based metric that evaluates the quality of generated text by comparing it to reference outputs using contextual embeddings rather than exact word overlap. Unlike traditional metrics like BLEU and ROUGE, which rely on n-gram matching, BERTScore leverages pre-trained transformer models (such as BERT) to compute token-level embeddings and measure their similarity via cosine similarity. This approach allows BERTScore to capture semantic meaning, making it particularly useful for evaluating RAG-generated responses where lexical variations do not necessarily imply a difference in intent. By going beyond surface-level string matching, BERTScore provides a nuanced assessment of how well a generated response aligns with its reference in terms of meaning and relevance.

Utilizing LLM-as-a-judge methods leverages models' robust semantic capabilities to compare generated responses against reference texts through dynamic reasoning processes \cite{snowflake2025benchmarking}, usually surpassing performance of classic methods and BERTScore (which tends to utilize small language models - SLMs).

Modern implementations employ three principal LLM-judge architectures for reference-based assessment:

\begin{itemize}
    \item \textbf{Cross-attention Scoring}: The judge LLM analyzes alignment between reference and generated text through attention mechanism analysis, particularly effective for technical documentation verification \cite{aws2025bedrock}.
    \item \textbf{Chain-of-Thought Comparison}: The model generates step-by-step reasoning comparing semantic equivalence between reference and response \cite{arize2025survey}.
    \item \textbf{Multi-Aspect Rubric Evaluation}: LLM judges score outputs against reference texts using predefined dimensional rubrics (e.g., factual consistency, stylistic match) \cite{confident2025judge}.
    
\end{itemize}

For instance, the Amazon Bedrock RAG Evaluation Service exemplifies production-grade implementation, combining reference-based LLM judging with automatic rubric generation \cite{aws2025bedrock}. Their three-phase process:

\begin{enumerate}
\item \textbf{Reference Analysis}: LLM extracts key claims and semantic structures from reference text
\item \textbf{Cross-Examination}: Generated response is checked against reference-derived constraints
\item \textbf{Divergence Scoring}: Model quantifies deviations using context-aware similarity measures
\end{enumerate}

This approach achieved 92\% precision in detecting factual inconsistencies in legal document analysis benchmarks, compared to 78\% for BERTScore-based evaluations \cite{aws2025bedrock}. The system's explainability features provide natural language rationales for each discrepancy, addressing a critical limitation of black-box neural metrics.

\subsubsection{Reference-free Evaluation}
\hfill \break

 While reference-based metrics, relying on human-annotated ground truth answers, remain valuable, they are often expensive to obtain and may not fully capture the nuances of RAG performance, particularly in open-domain scenarios. This has led to a surge of interest in \textit{reference-free} evaluation metrics. These metrics offer the advantage of assessing RAG outputs without requiring pre-defined gold standard answers.

Reference-free metrics can consider different combinations of input and output elements, such as assessing the relevance of the generated answer to the input query, evaluating whether the generated answer is grounded in the retrieved context, or measuring the overall quality and coherence of the generated text itself. In essence, reference-free metrics provide a practical and scalable way to evaluate RAG systems, offering valuable insights into different aspects of their performance.

Roughly speaking, there are three main types of reference-free metrics for evaluating the generation phase of a RAG system:

\textbf{1. Answer Relevance and Quality:} RAG evaluation assesses answer relevance and quality without requiring a ground truth. Metrics include sentence embedding similarity (e.g., Sentence-BERT \cite{reimers-gurevych-2019-sentence}) and LLM-based scoring for relevance, helpfulness and completeness.

\textbf{2. Context Groundedness and Faithfulness:} Answers should be grounded in retrieved context. Metrics evaluate factuality by checking if the answer is entailed by the context (e.g., using NLI models \cite{bowman-etal-2015-large, williams-etal-2018-broad}) and measure how much the answer relies on the retrieved content.

\textbf{3. Human Alignment and Safety:} Evaluating alignment ensures the system follows instructions and user intent, using compliance scoring and user feedback simulation (e.g., Pistis-RAG \cite{bai2024pistis}). Safety measures include bias detection, sentiment analysis, and content moderation to prevent harmful outputs.

\subsection{Holistic and Multi-Dimensional Evaluation Benchmarks}

Evaluating RAG systems requires holistic frameworks that integrate retrieval and generation assessment while ensuring scalability and task generalizability. Recent advancements have introduced unified, automated, and modular approaches to streamline evaluation across diverse applications.

RAGChecker \cite{ru2024ragchecker} and CoFE-RAG \cite{liu2024cofe} provide end-to-end evaluation by simultaneously analyzing retrieval effectiveness and generation fidelity. These frameworks offer diagnostic tools for transparency, error localization, and interpretability, enabling systematic debugging and optimization.

 RAGAS \cite{es2023ragas} minimizes reliance on ground-truth annotations by leveraging unsupervised, language model-based metrics. This approach facilitates scalable evaluation of retrieval relevance, faithfulness, and context alignment, making it adaptable to low-resource and dynamic domains.

By integrating multi-dimensional assessment methodologies, these frameworks establish a foundation for benchmarking RAG systems effectively, promoting reliability, reproducibility, and generalizability in diverse real-world applications.

\subsection{Domain Specific Evaluation Benchmarks}
While generic benchmarks offer initial assessments of RAG models, they often fail to capture real-world performance in specialized fields. Since RAG effectiveness depends on domain-specific information quality, developing targeted evaluation benchmarks is crucial for assessing models within particular subject areas, data types, and tasks.

For industry applications, domain-specific benchmarks complement, rather than replace, evaluations using datasets reflective of real-world deployment. Public benchmarks rarely match the data a system encounters in production, making customized evaluation datasets essential.

Several research efforts have addressed this need:

\begin{itemize}
    \item \textbf{Code Generation and Analysis:} Domain-specific evaluation is vital for RAG models handling code tasks like completion, search, bug fixing, and synthesis. CodeRAG-Bench \cite{wang2024coderagbenchretrievalaugmentcode} emphasizes evaluating models on code execution, syntax, and functional correctness, aspects often missing in generic text benchmarks.
    \item \textbf{Medical Applications:} RAG in healthcare requires high accuracy and reliability. Medical RAG benchmarks \cite{xiong2024benchmarkingretrievalaugmentedgenerationmedicine} assess clinical accuracy, factual correctness, and safety using medical knowledge bases, guidelines, and real-world patient data.
    \item \textbf{Legal Domain:} Legal RAG tasks, such as case law analysis and legal document retrieval, require benchmarks that assess legal reasoning, precedent identification, and accurate statutory interpretation \cite{pipitone2024legalbench}.
    \item \textbf{Financial Domain:} RAG models used for financial analysis and market predictions must be evaluated on financial datasets, considering accuracy, risk assessment, and regulatory adherence \cite{wang2024omnieval}.
\end{itemize}
Domain-specific benchmarks provide realistic performance assessments, enable targeted optimizations, and clarify RAG strengths and weaknesses across fields. Future work should focus on standardizing benchmark creation, developing domain-specific metrics, and fostering cross-domain collaboration to advance RAG evaluation.

\section{Deepchecks’ RAG Evaluation Framework}
Deepchecks is a modular end-to-end evaluation system for LLM-based applications, designed to deal with various use-cases, including RAG. Deepchecks’ framework incorporates core principles from leading RAG evaluation methodologies and provides an intuitive user-interface for evaluation and production monitoring.

Deepchecks evaluation framework decomposes complex tasks into smaller, manageable sub-tasks, referred to as "properties". Each property is evaluated independently, enabling error tracing and iterative improvements in pipeline performance. In addition to property-level assessments, Deepchecks employs a learned aggregation mechanism to derive an end-to-end evaluation metric. This feature allows for a holistic assessment of the pipeline’s performance while maintaining transparency in individual component evaluations.

The above framework supplies a solid foundation for evaluating the performance of a RAG pipeline. Deepchecks complements it with tools for root cause analysis (RCA), comparing performance of different system versions, and monitoring it once it's deployed in a production environment.

\subsection{Properties for RAG Evaluation}  
In the context of RAG evaluation, Deepchecks decomposes the task into three main properties, accompanied by several safety properties:
\begin{itemize}
    \item \textbf{Retrieval Relevance:} Measures that the retrieved information is relevant for addressing the input question.
    \item \textbf{Grounded in Context:} Measures that the generated answer is factually consistent with the source of truth, synonymous with faithfulness and adherence to the source of truth.
    \item \textbf{Completeness:} measures that the answer comprehensively addresses all parts of the question.
    \item \textbf{Safety Properties:} Include metrics for evaluating safety concerns such as toxicity, inclusion of personally identifiable information (PII), jailbreak attempts, and other ethical considerations. As in any LLM-based application, safety properties are essential metrics for RAG that ensure the LLM operates within ethical, legal, safe, and unbiased boundaries.
\end{itemize}

These principles form the foundation of several popular frameworks (\cite{es2023ragas}, \cite{ru2024ragchecker}), and Deepchecks combines them into an integrated system while excelling in accurately and efficiently measuring these properties. For each property, it utilizes state-of-the-art techniques alongside proprietary methods designed to perform exceptionally well in real-world scenarios. An example of these methods will be presented in section 4.

In cases where a reference answer that can serve as ground truth is available, Deepchecks provides the \textbf{Expected Output Similarity} property. This metric evaluates whether the generated output captures the main arguments present in the reference output, functioning as either a sufficient evaluation metric or a complementary measure to reference-free properties.

\subsection{End-to-end RAG Evaluation}  
In addition to property evaluation, Deepchecks emphasizes capturing human intuition by determining whether interactions are positive or negative, ensuring the evaluation framework aligns with real-world expectations and user requirements. Deepchecks' evaluation is the only framework to date to generate a holistic metric for RAG pipelines.

This approach enables Deepchecks to deliver a nuanced and accurate assessment of RAG pipelines, ensuring alignment with user expectations and practical use cases. A thorough analysis of Deepchecks’ holistic evaluation will be presented in section 5.

\subsection{Tools for Research and Production Life-Cycle}  
Deepchecks’ RAG evaluation capabilities extend beyond calculating metrics and generating end-to-end scores. By offering advanced tools for RCA, version comparison, and production monitoring, Deepchecks provides actionable insights that drive informed decision-making and system improvements. This section explores how these tools complement the metrics calculated by Deepchecks.

\subsubsection{Root Cause Analysis (RCA)}
Deepchecks’ RCA tools allow practitioners to identify the underlying factors contributing to performance degradations or anomalies in RAG systems. Using the detailed breakdown of properties, users can pinpoint specific components within the pipeline that require improvement. 

For example, RCA can help determine whether retrieval failures or inadequate grounding in the retrieved context are responsible for suboptimal responses. This targeted approach accelerates debugging and enables focused iteration on the weakest components of the system.

\begin{figure}[ht]
\centering
\includegraphics[width=0.8\textwidth]{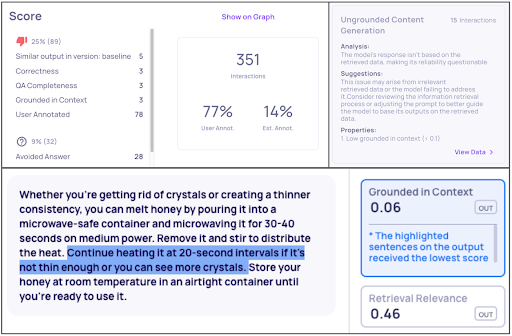}
\caption{: Deepchecks’ RCA tools, such as annotation breakdown (top-left), insights based on properties’ scores (top-right) and ungrounded content highlighting (bottom) assist in pinpointing specific components within the pipeline that require improvement.}
\label{fig:rca-tools}
\end{figure}
\subsubsection{Version Comparison}
The version comparison feature is a critical tool for assessing the impact of changes made to a RAG system. By juxtaposing the metrics of different system versions side-by-side, users can evaluate whether modifications have led to improvements in key properties. 

Deepchecks allows for both per-metric comparisons (e.g., the effect of a new retrieval model on context relevance) and per-interaction analyses (e.g., comparing outputs for specific queries). This structured comparison helps users understand trade-offs and ensures that system upgrades align with desired objectives.

\subsubsection{Production Monitoring}
Once deployed, a RAG system must maintain its performance under varying real-world conditions. Deepchecks’ production monitoring tools track key reference-free metrics over time, enabling users to detect and respond to performance drifts or anomalous behavior. 

\begin{figure}[ht]
\centering
\includegraphics[width=0.8\textwidth]{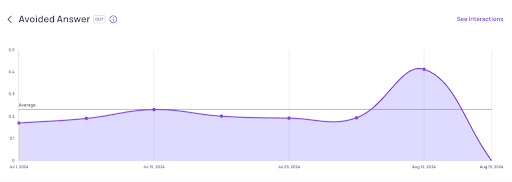}
\caption{: Deepchecks’ production monitoring tracks key metrics over time, indicating performance degradation due to data distribution shifts and assists in prompting focused system upgrades.}
\label{fig:metrics-over-time}
\end{figure}
For instance, an increase of answer avoidance score over time likely indicates data distribution shifts. By continuously evaluating system metrics, organizations can ensure their RAG systems remain reliable and effective, even as external conditions evolve.

\section{Grounded in Context}
As mentioned in the introduction, one of the most pressing challenges in generative AI is addressing hallucinations—cases where a language model produces inaccurate or unsupported claims. Researchers are actively exploring a variety of approaches for hallucination detection, aiming to enhance the reliability of LLM-generated content. Table~\ref{tab:faithfulness-comparison} compares the factual grounding faithfulness of Deepchecks, RAGAS, and Langsmith across several datasets.

\begin{figure}[ht]
\centering
\includegraphics[width=0.8\textwidth]{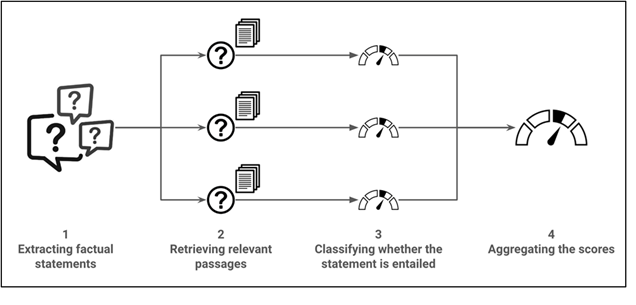}
\caption{: Deepchecks’ Grounded in Context method. Calculating an entailment score for each factual statement based on the chunks of the documents most relevant for it, and then aggregating the scores to a unified Grounded in Context score.}
\label{fig:grounded-pipeline}
\end{figure}

RAGAS is an open-source evaluation framework for RAG applications, providing metrics to assess faithfulness, relevance, and retrieval quality. It integrates with popular LLMs to diagnose and improve RAG system performance. LangSmith is a LangChain tool for debugging, monitoring, and evaluating LLM applications, offering insights into prompt execution, latency, and model behavior. It enables developers to trace and optimize their pipelines with structured logging and performance analytics. 

Both RAGAS and LangSmith rely solely on external LLMs to evaluate various metrics, using their own predefined prompts for the process.

Deepchecks’ property for evaluating the faithfulness of a generated output to a source of truth is called \textit{Grounded in Context}, and it is comprised of a multi-step pipeline (see Figure~\ref{fig:grounded-pipeline}):

\begin{enumerate}
    \item Extracting factual statements from the LLM's output using a proprietary small language model (SLM) for information density classification.
    \item Retrieving relevant passages from the context for each statement.
    \item Classifying whether a statement is entailed by the retrieved passages using a proprietary SLM trained on production-like data.
    \item Aggregating the entailment scores using a dedicated averaging method.
\end{enumerate}

\subsection{Benchmarking Grounded in Context}
To benchmark this method, we compared its performance against other established approaches. The datasets used in this comparison included context as the source of truth, model outputs, and annotations indicating whether the outputs were factually grounded. The classification setup required each method to score the factual grounding of the outputs, with results evaluated using the ROC-AUC metric.

\begin{table}[ht]
\centering
\begin{tabular}{|l|c|c|c|}
\hline
\textbf{Dataset}
& \multicolumn{1}{|p{2cm}|}{\centering Deepchecks Grounded \\ in Context}
& \multicolumn{1}{|p{2cm}|}{\centering RAGAS Faithfulness \\ (GPT-4o)}
& \multicolumn{1}{|p{2cm}|}{\centering Langsmith Answer \\ Faithfulness (GPT-4o)}\\
\hline
TRUE & \textbf{0.86}& 0.65 & 0.63 \\
SQuAD & \textbf{0.92}& 0.83 & 0.65 \\
PubmedQA & \textbf{0.84}& 0.80 & 0.78 \\
\hline
\end{tabular}
\caption{Performance of Deepchecks, RAGAS, and Langsmith faithfulness measurement methods in evaluating factual grounding across several datasets.}
\label{tab:faithfulness-comparison}
\end{table}

The following datasets were used in the evaluation:

\begin{itemize}
    \item \textbf{TRUE Benchmark:} A comprehensive collection of 11 manually annotated datasets from various domains. We sampled 100 entries from each dataset.
    \item \textbf{Corrupted Open Datasets:} GPT-4 was used to generate two types of negatives—hallucinations and contradictions—based on the SQuAD dataset. Each dataset contained 62 positive samples, 31 with hallucinations, and 31 with contradictions.
    \item \textbf{PubmedQA}: a biomedical question answering (QA) dataset collected from PubMed abstracts. We sampled 1000 entries.
\end{itemize}

\section{RAG End-to-End Evaluation}

Deepchecks' end-to-end (E2E) evaluation integrates the scores from the core properties—retrieval relevance, grounded in context, completeness, and safety—through a learned aggregation mechanism. This mechanism classifies each triplet (question, retrieved-context, and answer) into one of three categories:
\begin{itemize}
    \item \textbf{Negative}: A single failure in any of the core properties automatically classifies the interaction as negative.
    \item \textbf{Positive}: To qualify as positive, the answer must meet a high threshold across all core properties, ensuring robustness and reliability.
    \item \textbf{Unknown}: Interactions that don’t fail any metrics but also fall short of the stringent criteria for positive classification are labeled as unknown.
\end{itemize}
While Deepchecks provides pre-configured settings for comprehensive RAG evaluation, the platform allows for customization of threshold values across the core metrics—groundedness, relevance, completeness, and safety—as well as incorporation of additional metrics to align with specific evaluation requirements.

\subsection{Benchmarking End-to-End Evaluation}
To assess the quality of Deepchecks' automatic evaluation pipeline, we collected various annotated Retrieval-Augmented Generation (RAG) datasets and fed the pipeline with triplets of question, context, and answer. A positive triplet is defined as one where the answer is both correct and faithful (i.e., grounded in the provided context). Deepchecks' evaluation of the answers was then compared with human-provided annotations.

We compared our results with two alternative LLM evaluation frameworks: Langsmith and RAGAS. Neither framework offers an evaluation pipeline that provides a general annotation (positive/negative). Thus, we aggregated their supplied metrics in a way that yields the best possible score for each. To ensure a fair comparison, we used only the property-based evaluation component of Deepchecks' pipeline, as competitors’ systems do not learn from client data. This approach facilitates a balanced assessment across all systems while underrepresenting the full capabilities of Deepchecks' complete pipeline.

Each dataset was balanced between positive and negative examples, and performance was evaluated using the accuracy metric (\textit{number of correct predictions / total number of samples}). Accuracy is most suitable for balanced binary classification tasks where both labels are of interest.

\subsubsection{\textbf{Client Datasets}}
\begin{itemize}
    \item Real-world RAG use cases from two of our clients, featured in this demo with their consent.
    \item Provided with client annotations based on specific needs.
    \item Most representative of production use cases.
    \item Most challenging to analyze correctly.
    \item Each dataset contains 100 samples with balanced binary labels (annotated by the clients).
    \item Derived from two use cases: Q\&A on technical support data and job candidates' work histories.
\end{itemize}

\subsubsection{\noindent \textbf{Public Datasets}}
\begin{itemize}
    \item Widely known and reliable public RAG use-case datasets: \texttt{SQUAD}, \texttt{RAG-dataset-12000}, and \texttt{HAGRID}.
    \item Originally contained only positive samples; negatives were generated using several corruption methods:
    
    \begin{itemize}
        \item \textbf{Contradiction:} Partially modifying the answer to create internal inconsistencies.
        \item \textbf{Hallucination:} Altering facts in the answer to produce outputs that are not fully grounded in the given context.
        \item \textbf{Irrelevant (but faithful):} Replacing the answer and context with similar but unrelated information (most similar negatives in the dataset).
    \end{itemize}
    \item Each dataset contains 100 samples with balanced binary labels (original samples annotated as positive; corrupted samples annotated as negative).
\end{itemize}

\subsubsection{Results}

The results of the evaluation are summarized in Table~\ref{tab:results-comparison}. Deepchecks demonstrates superior or competitive performance compared to Langsmith and RAGAS across all datasets, particularly excelling in client datasets representative of real-world production use cases.

\begin{table}[ht]
\centering
\begin{tabular}{lccc}
\hline
\textbf{Dataset} & \textbf{Deepchecks} & \textbf{RAGAS (GPT-4o)} & \textbf{Langsmith (GPT-4o)} \\
\hline
\texttt{RAG-dataset-12000} & \textbf{0.96}& 0.65 & \textbf{0.96}\\
\texttt{SQUAD}             & \textbf{0.94}& 0.78 & 0.93 \\
\texttt{HAGRID}            & \textbf{0.94}& 0.59 & 0.84 \\
\texttt{Client 1 (Tech Support)}          & \textbf{0.70}& 0.55 & 0.58 \\
\texttt{Client 2 (Job Profiles)}          & \textbf{0.81}& 0.46 & 0.54 \\
\hline
\end{tabular}
\caption{Performance comparison of Deepchecks, RAGAS, and Langsmith across public and client datasets, measured by accuracy.}
\label{tab:results-comparison}
\end{table}

\section{Conclusions}

In this study, we presented a comprehensive evaluation of Deepchecks, a modular and end-to-end system for assessing the quality of Retrieval-Augmented Generation (RAG) pipelines. By focusing on core properties such as retrieval relevance, contextual grounding, completeness, and safety, Deepchecks provides a robust framework for diagnosing and improving the performance of RAG-based systems. 

Through detailed benchmarking against alternative solutions like RAGAS and Langsmith, we demonstrated that Deepchecks outperforms or matches existing methods across both public and client datasets. Specifically, Deepchecks excels in scenarios involving real-world, production-level complexity, as evidenced by its higher accuracy on client-provided data. This highlights the framework's ability to meet practical demands and align with user expectations in diverse deployment contexts.

Our results further underscore the importance of property-based evaluations in mitigating common challenges associated with LLMs, such as hallucinations, contradictions, and irrelevant outputs. By incorporating a learned aggregation mechanism for end-to-end assessments, Deepchecks is uniquely positioned to provide holistic evaluations that capture both granular and global quality measures.

In conclusion, Deepchecks framework represents a significant step forward in the evaluation of LLM-driven RAG systems, offering a reliable, accurate, and user-aligned tool for advancing the state of the art in generative AI applications.

%
%
%
\bibliographystyle{splncs04}
\bibliography{mybibliography}

\end{document}